# Extending an Information Extraction Tool Set to Central and Eastern European Languages

**Camelia Ignat, Bruno Pouliquen, António Ribeiro & Ralf Steinberger**
European Commission - Joint Research Centre
Institute for the Protection and Security of the Citizen (IPSC)
T.P. 267, 21020 Ispra (VA), Italy
http://www.jrc.it/langtech
`Firstname.Lastname@jrc.it`

## Abstract

In a highly multilingual and multicultural environment such as in the European Commission with soon over twenty official languages, there is an urgent need for text analysis tools that use minimal linguistic knowledge so that they can be adapted to many languages without much human effort. We are presenting two such Information Extraction tools that have already been adapted to various Western and Eastern European languages: one for the recognition of date expressions in text, and one for the detection of geographical place names and the visualisation of the results in geographical maps. An evaluation of the performance has produced very satisfying results.

## 1 Introduction

The international staff of the European Commission (EC), like any other multinational organisation, has to deal with documents written in many different languages. Multilingual text analysis tools can help them to be more efficient and to get access to information written in documents they may not understand. However, not many commercial text analysis tools exist that can analyse texts in all official European Union (EU) languages, and we do not know of any tool that covers all of the over 20 languages that will be used after the planned Enlargement of the EU. The *Joint Research Centre* (JRC) of the EC therefore uses and develops text analysis tools that require very few linguistic resources so that algorithms developed for one language can easily be adapted to the other languages. The adaptation of existing tools to the languages of the Central and Eastern European countries soon joining the EU has become a priority for the JRC.

In this paper, we present two information extraction (IE) tools that can easily be adapted to new languages: the recognition of date expressions (which can easily be adapted to the recognition of currency expressions) and the recognition and visualisation of geographical place names in text. Our platform-independent tools are implemented using object-oriented PERL, currently interfacing with Oracle, but any relational database could be used. The CGI interface can be viewed with any conventional web browser.

For an introduction to the state of the art of the field of Named Entity Recognition (NER), see Daille & Morin (2000). For an alternative system for multilingual NER, see Pastra et al. (2002).

Section 2 gives a brief overview of the JRC's tool set. Section 3 discusses the difficulties having to do with the different character sets used in the European and other languages. Sections 4 and 5 present the tools to recognise place names and date expressions. The conclusion summarises and points to planned work.

## 2 JRC's IDoRA system

We refer to the collection of JRC tools to gather and analyse documents and to visualise the contents of individual texts or of whole document collections as IDoRA (*Intelligent Document Retrieval and Analysis*). The **gathering component** consists of a crawler with parameter files that tell it where to start its search, where not to go, which search word combinations to look for, what sort of file types to download, etc. (Steinberger et al. 2003).

The **analysis component** includes document format converters, a statistical language and character encoding recognition tool trained for 25 languages, tools to identify keywords and controlled vocabulary descriptors (Pouliquen et al. 2003), tools to extract date expressions and references to geographical place names, software to produce subject-specific summaries and tools to cluster and classify documents, as well as to calculate document similarity. Several applications not only work for several languages (multi-monolinguality), but are also cross-lingual, meaning that the contents of texts in different languages can be compared to each other or that the contents of one language can be presented in another.

The **visualisation component** provides means to search and visualise the information extracted from one or more documents in a document profile or in a geographical map.

The two tools presented here are clearly part of the analysis component. Like most other JRC tools, they are mainly based on statistics and heuristics, with a minimal linguistic input and with no part-of-speech tagging or grammar. We believe that adding linguistic rules would improve the results, but developing these rules for so many languages is out of our reach. Instead, we try to optimise tools using this knowledge-poor approach.

## 3   Character Encoding

Character encoding handles the way characters are converted into bytes – the character set. For example, 'A' is encoded by the sequence of bits '01000001' in ASCII.

We experienced several problems with character encoding when we started to work with Central and Eastern European languages. For example, Hungarian texts cannot be encoded with ISO-8859-1, which is the encoding used for most Western European languages. It needs some characters not available in this character set, like the character 'ő' as in 'bővítés' (*enlargement*).

Unlike English, which is able to encode its characters with the 128 possible characters of the *American Standard Code for Information Interchange* (ASCII), the problem with character encoding started early on when the Western European languages required the usage of characters with diacritics, like 'à' in French or 'ü' in German.

To cope with this, the ASCII character set was extended and this led to the creation of the ISO-8859 family of character sets which is a standard of the *International Standards Organisation* (ISO). For example, the ISO-8859-1 character set covers most Western European languages, the ISO-8859-2 (also known as Latin 2) covers Central and Eastern European languages, ISO-8859-3 covers Turkish and Maltese, ISO-8859-5 covers Bulgarian and other Cyrillic languages, and the ISO-8859-7 character set covers the characters needed for Greek. In this way, it became possible to write in any of these character sets, but it was not possible to mix different character sets in the same file.

Both ISO and Unicode were initially and independently developing a universal character set to make it possible to save and read files with characters in different languages. The *Universal Character Set* standard (UCS) is supposed to cover virtually all known languages and it is defined in an agreement between ISO and Unicode. UTF-8 (*UCS Transformation Format-8*), another ISO standard, encodes each character as a sequence of one or more 8 bit bytes. E.g. the character 'A' is encoded as in a single byte '65' but the character 'ü' is represented by a sequence of two bytes '195 188' (in Big-Endian convention, i.e. the most significant byte comes first). UTF-8 was created because some systems were sensitive to characters with special meanings, such as control characters, and also to provide backward compatibility with ASCII. It has the virtue that a pure ASCII file is also a valid UTF-8 file.

Bearing in mind the various existing encodings, it became clear that it was important to handle character encoding identification. So, for each language, we collected texts encoded in the different encodings, to build a typical 'profile' for each language-encoding pair. We used an algorithm for language recognition based on Markov Models by Ted Dunning (1994). We start by compiling a list of frequencies of bi- and trigrams of bytes for each training text for a specific language encoding. We do the same for the text whose language encoding we want to identify and apply the algorithm to compute a similarity score with the various language encodings and assign the language encoding with highest score. This means that the recognition of the character set and of the language is done in the same step. According to Ted Dunning (1994, p. 16) this algorithm has a high precision even for small text inputs and small training data sizes, and our tool has indeed always identified the correct language-character set combination.

We should stress that the development of multilingual applications does involve various prob-lems. For instance, Oracle 8i databases should be configured to UTF-8 so as to support the storage of characters in various languages; JAVA uses natively the UCS-2 character encoding; and, PERL has only recently moved to support natively UTF-8, in version 5.8. Any combination of these tools and languages requires great care.

## 4 Recognition and Visualisation of Place Names

According to Friburger & Maurel (2002), 43.9% of proper names are locations. Gey (2000) furthermore claims that 30% of content-bearing words in journalistic text are proper names. They both showed, respectively, that clustering and Information Retrieval can be improved when considering proper names. Software to identify proper names is therefore an important part of a language technology tool set.

The tool presented here takes plain text files as input and produces as output lists of place names referred to in the text. Furthermore, it links the place names found with further database information on these places, namely the countries they belong to, their geographical coordinates and their size classification category (from 1 = capital and 2 = major city to 6 = village). For each document, or collection of documents, the tool can then calculate the total number of hits per country and the percentage of hits per country, as compared to all hits. The geographical coordinates can be used to produce maps such as those in **Figures 1 and 2**, in which the countries and the cities are highlighted according to their relative frequency with which they are mentioned (viewing in colour is recommended).

The identification of place names always requires name lists and cannot be done with linguistic patterns because names of locations have few reliable contextual clues (Mikheev et al., 1999). We use the commercial *Global Discovery* database from *Europa Technologies Ltd*, UK, which contains over half a million place names world-wide. For the visualisation of the geographical maps, we use the *Digital Map Archive* tool developed in the context of the JRC's IS-

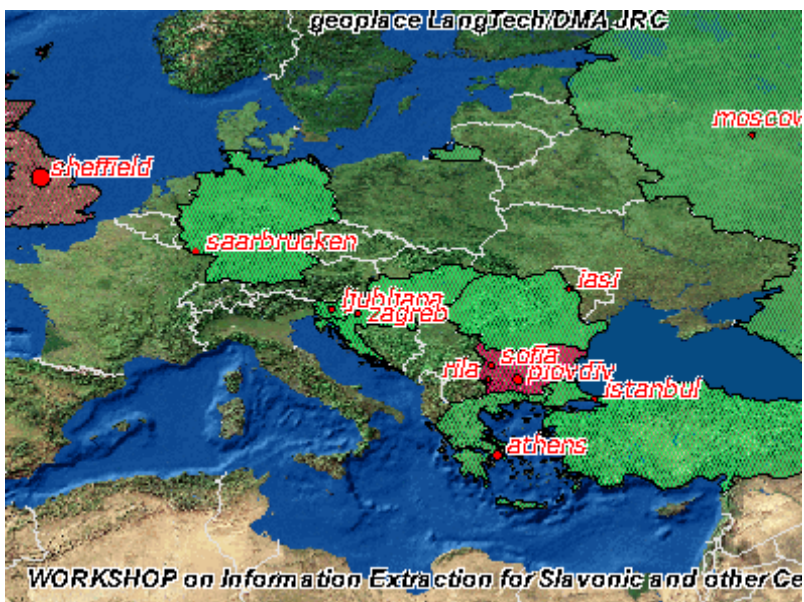

**Figure 1.** Geographical visualisation of the place names contained in the text of the RANLP conference web site http://lml.bas.bg/ranlp2003/. Cities are highlighted by dots of varying sizes and countries by various colour codes, depending on how often they were referred to in the text.

FEREA project (Ehrlich et al., 2003; http://dma.jrc.it/). It provides map generation tools with multiple layers to add information on roads, airports, lakes, population density and more. It also contains satellite images of the world and allows a very realistic view of the area displayed. For a list of gazetteers freely available on the internet, see Gey (2000).

The **difficulties** of the task are linked to (a) multiple places that share the same name, such as the fourteen cities and villages in the world called 'Paris', (b) place names that are also words in one or more languages, such as 'And' (Iran), 'Split' (Croatia) and 'Annan' (UK), and (c) places that have varying names in different or even in the same language (Italian 'Venezia' vs. English 'Venice', German 'Venedig', French 'Venise'; 'Leningrad' vs. 'Saint Petersburg', etc.). The latter problem can be solved by completing the database with translations and variations. A good list of name variations can be found, for instance, at the KNAB database website of the *Institute of the Estonian Language* (see http://www.eki.ee/index.html.en).

The JRC tool relies on a simple **dictionary lookup** in the text and does not currently check any local patterns, which makes the system most language-independent. In addition to the place name list, we added lists of country ISO codes, of currency names, of adjectives pointing to a country or to the people of a country. This means that a hit for the country is generated even if only its currency or its people (e.g. 'Iraqi') are mentioned.

In the languages we are currently dealing with, place names are always written in upper case. The tool therefore parses the text and checks for each upper case word whether it is either a place name by itself or whether it is the beginning of a multi-word place name (e.g. 'Stara Zagora' in Bulgaria). In the latter case, a regular expression checks whether the right-hand-side context contains the remainder of the multi-word name and distinguishes, for instance, 'Stara Zagora' in Bulgaria from similar place names such as 'Stara Lubovna' (SK), 'Stara Pazova' (YU), 'Stara Reka' (BG), 'Stara Tura' (SK), 'Stara Wrona' (PL) and others.

In order to avoid common words such as 'And' and 'Annan' from triggering the tool, we use language-specific '**geo stop word' lists** containing strings with place name homographs that the system ignores when found in text of that language. These lists were generated, separately for each language, by matching all geographical place names against the most frequent words of that language, and by hand-selecting the matches (to avoid place names like 'London' and 'Paris' to be in the stop word list). For English, more 'place stop words' were added later after an error analysis of the most frequent hits in a newspaper text corpus of 250,000 articles.

We reduced the half million place names in the database to about 85,000 by taking out the smaller places outside Europe, i.e. those with a size class of three to six. The consequent lower recall is acceptable because the majority of our text sources is from Europe and the US and tends to mention the country name when talking about smaller places outside Europe, which means that the fact that there is a reference to a specific country would still be identified. This reduction improved the computational efficiency and reduced noise.

For the reference **disambiguation** of place names used for several cities (e.g. 'Roma' in Italy and 'Roma' in Romania), we use heuristics that mainly use two types of information: size classification and country. By default, the place with the larger size code is used (the capital of Italy),

**Armistitiu**
In fiecare an, la *11 noiembrie*, in *Franta* si in *Marea Britanie* se celebreaza Armistitiul prin care s-a pus capat primului razboi mondial. In *Franta* este chiar zi oficiala de sarbatoare, jour férié, nu se lucreaza; in *Marea Britanie*, comemorarea are loc in cea mai apropiata duminica de data de 11 (se si cheama Remembrance Sunday, Duminica Aducerii-aminte, s-a tinut alaltaieri).
Nu se sarbatoreste nimic insa in *Germania*, ceea ce pina la un punct este explicabil. Armistitiul, care a fost semnat la *11 noiembrie 1918* intr-un vagon-restaurant prefacut cu acel prilej in sala de reuniune, a parafat capitularea *Germaniei*. Acel vagon, aflat pe o linie secundara a garii satului *Rethondes*, in padurea linga orasul *Compiègne*, avea sa fie simbolic folosit si in 1940, de aceasta data pentru semnarea armistitiului care consfintea infringerea *Frantei* in fata *Germaniei* lui Hitler.

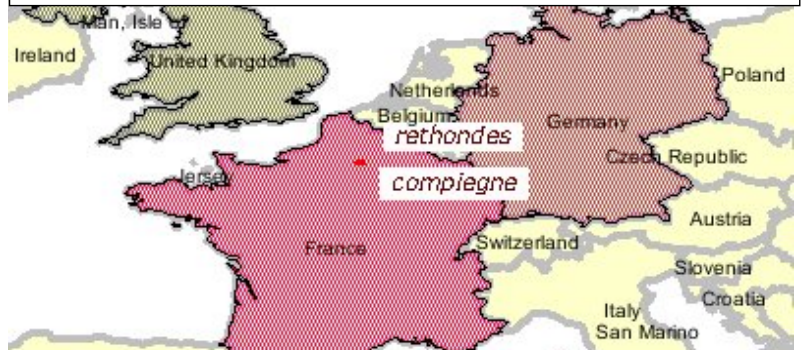

**Figure 2.** Example of Romanian date and place recognition.

but if the text contains more references for the country with the smaller place of the same name (e.g. 'Roma' in Romania), the latter is chosen. The disambiguation heuristics can still be improved. For instance, it may be useful to consider the source country of the analysed document. In a local Romanian newspaper article from the region around 'Roma', the local town should probably be chosen over the Italian capital.

A **manual evaluation** of 80 English texts (730 KB) containing a total of 853 references to geographical place names yielded a precision of 96.8% and a recall of 96.5%. Error analysis showed that the reason for missing entries (lowering recall) was that the database was not complete. For instance, some regions' (*Montenegro*; *Nord-Pas de Calais*) and peoples' names (*Timorese*; *Bosnian*) were not in our place name list and we have no names of mountains, rivers, islands, stretches of Sea, etc. (*Azores*; *Adriatic Sea*). On the other hand, precision was lowered when persons' names mentioned in the text had geographical homographs. For instance, the surname *Cox* triggered a city in France and the Nigerian General *Abasha* triggered a town with the same name in Georgia. Finally, the country *Malta* was wrongly disambiguated as a Portuguese town. As it is not clear that groups of states such as *EU*, *ACP*, *Benelux* and *Scandinavian Countries* are geographical places, we did not consider them in the evaluation.

The **limitation** of the current tool is that all place names exist only in the local language and in English. This means that translations of place names in languages other than English are not recognised. E.g. 'Venezia' and 'Venice' will be recognised, but not 'Venedig' and 'Venise'. We have access to translations of country names and capitals in the current eleven official EU languages and will add them, plus information collected from other sources, to our database.

Another limitation is that of the alphabet. All place names in our database are currently written with the Roman alphabet so that the recognition will work for alphabetical languages such as Romanian, but not, for instance, for Cyrillic languages such as Bulgarian. As our system is using UTF-8 encoding (see section 3), the tool could include place names written in Cyrillic letters. It is planned to include the historical, geographical, as well as alphabetic (Roman, Cyrillic, etc.) variations of place names from the KNAB database (see above, in the same section), after which it will be easier to evaluate the system on texts in further languages.

## 5 Recognition of Date Expressions

The extraction of date expressions is of interest if users search for documents talking about an event that happened in a certain period of time. Once date expressions have been identified and stored in a database, the database can be queried for documents mentioning a date in a certain time period. Furthermore, dates and other named entities can be used to improve the performance in tasks such as topic tracking, document similarity calculation or plagiarism detection.

Date expressions vary much in their format, depending on the style of a specific language, country or person. The date expression formats recognised by our system are shown in **Table 1.** Such expressions are recognised by our system, transformed into a *normal form*, and stored in the database, together with their offset and their length so that they can be found again and highlighted in the text, if necessary (see **Figure 2**).

While the task description in MUC conferences[1] is to tag *all* date and time expressions, including strings such as 'last Tuesday', 'next Summer', '4:15 PM', etc., we are not detecting such expressions because our clients are not interested in them. Should this need arise in the future, this functionality could be added.

Dates are often **incomplete** (e.g. '3$^{rd}$ May' or 'May 2001'). Therefore, we represent dates as one of three possible *types*: full date DDMM-YYYY with day (D), month (M) and year (Y); date without the day (MMYYYY) and date without the year (DDMM). In addition to these, relative dates such as 'yesterday' and 'last May' are recognised and can be resolved by making reference to the day of writing or publication of the document (the *reference date*; currently not implemented). For newspaper articles, for instance, the reference date is the publication date.

Table 1 shows that we do not currently recognise *periods* of time ('7-9 September'), seasonal expressions ('Summer 2003') and incomplete dates showing only the year ('in 2003'). These can be added easily when the need arises, but cases such as 'in 2003' are prone to produce many wrong hits.

The tool consists of PERL code and one **parameter file** per language containing the language-specific information. This includes the names and abbreviations of the days, months and relative time expressions and a list of pre- and post-modifiers such as 'last' in 'last May'. It also includes a list of words that can be found in between the parts of the date expression (e.g. '3$^{rd}$ *of* May', French 'trois mai *de l'année* 2003', Romanian 'întîi martie *al anului* 1998'). For languages where texts can be found with and without diacritics (e.g. French titles, or Romanian online texts), the accentuated and the non-accentuated vocabulary is added to the parameter file so that these variants can be recognised. Furthermore, in inflected languages such as German and Romanian, the possible case (or other) inflections can be added to the parameter file, as well.

Currently, parameter files exist and have been tested for the seven languages English, French, German, Italian, Portuguese, Romanian and Spanish. It is now planned to work on Czech, Polish and Hungarian. Creating a new language parameter file is only half a day's work, but finding an appropriate test corpus and testing the results obviously takes a bit longer.

The date recognition tool first detects **complete date expressions** in numerical form (e.g. '13/02/03' and '31.5.2003') by matching the text against a regular expression and extracts the items found. Next, it searches for full or abbreviated month names in the text. If the language is not known, we first apply our statistical language identification tool, which is trained for 25 languages. Each time a month name is found, the context to both sides is searched for the rest of the date. If successful, the date expression is extracted and stored in the database, together with offset, length and normalised date format.

---

[1] See, for instance, the MUC-7 specification at http://www.itl.nist.gov/iad/894.02/related_projects/muc/proceedings/ne_task.html

| Type | Description | Dates recognised, e.g. | Dates not recognised, e.g. |
|---|---|---|---|
| Complete absolute dates | Numerical; Containing month name (full or abbreviated) | *3-04-03* or *21.2.1983* or *1997/04/01* *1999, the 2nd of May* *the sixth of March in the year nineteen eighty four* | *1.2.15* *7 May 2003* in the period expression *7-8 May 2003* |
| Incomplete absolute dates | Containing month name (full or abbreviated) | *third February* *Jan. 2003* | incomplete numerical dates: *1990*; *the 1970s*; *two thousand and two* simple month name: *in May* |
| Relative dates | Relative to a reference day; Relative month names; Month name + relative year | *yesterday, today, tomorrow* *next June, last September* *February last year* | *last month; next Summer; Labour Day; on Tuesday; in the third quarter; February three years ago* |

**Table 1.** Description of date expression formats recognised by the JRC system.

Dates are often ambiguous and numerical expressions exist that look like dates but are not. The following **disambiguation** step tries to solve these ambiguities and unclear cases. The major ambiguity has to do with the American date format MDY, which contrasts with DMY, i.e. the inversion of the month and the day. This ambiguity is resolved by checking the number range of what is expected for the day and the month because there are only 12 months per year while there are up to 31 days per month. This helps to identify strings such as '12/31/03' as being MDY. For strings such as '01/02/03', where this method does not help, the system looks at further dates in the text to detect which date format is the standard of this document. If strings such as '12/31/03' are found, the date format MDY is detected and the date '01/02/03' is disambiguated as '2ⁿᵈ of January'. If no further dates are found so that no document standard can be established, we assume the default format DMY. The default format can be changed via a parameter.

We **evaluated** the tool on English and Romanian texts. The **English** test set consisted of 510 KB of texts (87,000 words) marked up with MUC-codes. An analysis of the test corpus showed that over 70% of time expressions are incomplete dates mentioning only the year or only a weekday without a date. As we are not trying to identify these, we simply ignored them in the evaluation. The evaluation yielded precision and recall values for relative dates of 86%/67%, for complete dates of 100%/100% and for incomplete dates of 98%/98%.

A string wrongly identified as a relative date was 'this May' in "this may sound". Relative dates not found were of the type 'late January' and 'mid-August'. We have since repaired this weakness by adding the strings 'late' and 'mid' as pre-modifiers to the parameter file so that the tool now identifies these incomplete dates.

One error lowering the precision for incomplete dates was due to an organisation name containing a date expression, which is an arguable mistake. In period expressions such as "7-8 June", we currently only detect the incomplete date '8 June' and miss '7 June'. We counted this as missing one date in the recall measure.

The **Romanian** test set consisted of 582 news articles from the Romanian newspaper *Evenimentul zilei*. It amounts to 1.6 MB of text and contains 1031 date expressions of the types that the system is supposed to recognise. The precision and recall achieved on this set are 97.7% and 98%. However, the real recall value may be a bit lower because it is possible that some date expressions were overlooked in the human evaluation.

Most of those date expressions not found by our system (lowering recall) were the first part of date range expressions (e.g. *intre 12 si 26 iunie*). The most common error lowering precision were due to the ambiguous string *mai*, which can be the month name *May* or the adverb *again/more*. For instance, in the string *cei doi mai incercasera*, the date 02/05 was erroneously identified.

The performance of the tool is now very good for those time expressions which we try to recognise. It would perform rather badly at a date recognition task according to MUC specifications because we are not trying to identify the large bulk of time expressions (weekdays or years without a date). The functionality of recognising these could, of course, easily be added. We would expect very high precision and recall for weekdays and slightly lower precision for years.

We are happy to give this tool to interested parties and would welcome help to extend the toolset to further, especially Central and Eastern European languages.

## 6 Conclusion

We have presented two software tools – one to detect and visualize geographical place names and one to detect date expressions – that can easily be adapted to new languages because the programming code is language-independent, no linguistic resources such as parsers are used and the language-specific information in the parameter files can be created or imported quickly and easily. We have pointed out that multilingual tools like these need software to detect the character set in which a text is written and that it is necessary to work with a character set such as UTF-8. Both applications work very well, with precision and recall rates over 95%. Better results can be achieved at the cost of adding more language-specific rules, which would make the tools more difficult to adapt to new languages.

Regarding place name recognition, future work will include an improvement of the disambiguation heuristics and the inclusion of more place name variations and of place names using Cyrillic and Greek characters. Regarding the date recognition tool, we will add parameter files for more (especially Central and Eastern European) languages. Furthermore, we will adapt the date recognition software to identify currency expressions.

## 7 Acknowledgement

We would like to thank Tom de Groeve from the JRC's ISFEREA project for his help, for providing us with geographical data from the *Digital Map Atlas* (http://dma.jrc.it, Ehrlich et al., 2003), and for giving us access to their powerful map generation system to create maps on the basis of the geographical references detected in text. We also thank Rada Mihalcea for providing us with a Romanian corpus, and the three anonymous reviewers for their suggestions and comments.